\pdfoutput=1

\documentclass[11pt]{article}

\usepackage{EMNLP2022}

\usepackage{times}
\usepackage{latexsym}

\usepackage[T1]{fontenc}

\usepackage[utf8]{inputenc}

\usepackage{microtype}

\usepackage{inconsolata}

\usepackage{xcolor, graphicx}
\usepackage[export]{adjustbox}
\usepackage{subcaption}
\usepackage{multirow}
\usepackage{amsfonts}
\usepackage{amsmath}
\usepackage{enumitem}
\usepackage{booktabs}
\usepackage{pifont}
\newcommand{\cmark}{\ding{51}}%
\newcommand{\xmark}{\ding{55}}%
\usepackage[normalem]{ulem}
\usepackage{hyperref,tablefootnote,footnotehyper}

\newcommand{\model}{ECT-BPS}

\title{ECTSum: A New Benchmark Dataset For Bullet Point Summarization of Long Earnings Call Transcripts}

\author{Rajdeep Mukherjee\textsuperscript{1}\thanks{$\;$ Corresponding author: rajdeep1989@iitkgp.ac.in.} $\quad$ Abhinav Bohra\textsuperscript{1} $\quad$ Akash Banerjee\textsuperscript{1} $\quad$ Soumya Sharma\textsuperscript{1} \\ {\bf Manjunath Hegde\textsuperscript{2} $\quad$ Afreen Shaikh\textsuperscript{2} $\quad$ Shivani Shrivastava\textsuperscript{2} $\quad$ Koustuv Dasgupta\textsuperscript{2}}\thanks{$\;$ Mentors' names are presented in alphabetical order.} \\ {\bf Niloy Ganguly\textsuperscript{1,3}\footnotemark[2] $\quad$ Saptarshi Ghosh\textsuperscript{1}\footnotemark[2] $\quad$ Pawan Goyal\textsuperscript{1}}\footnotemark[2] \\
        \textsuperscript{1} Department of Computer Science and Engineering, IIT Kharagpur, India \\ \textsuperscript{2} Goldman Sachs Data Science and Machine Learning Group, India \\ \textsuperscript{3} Leibniz University of Hannover, Germany}

\begin{document}
\maketitle
\begin{abstract}

Despite tremendous progress in automatic summarization, state-of-the-art methods are predominantly trained to excel in summarizing short newswire articles, or documents with strong layout biases such as scientific articles or govt reports.
Efficient techniques to summarize financial documents, including facts and figures, have largely been unexplored, majorly due to the unavailability of suitable datasets.
Here, we present \textbf{ECTSum}, a new dataset with \textit{transcripts of earnings calls} (ECTs), hosted by public companies, as documents, and short experts-written \textit{telegram-style bullet point} summaries derived from corresponding \textit{\textbf{Reuters}} articles.
ECTs are long unstructured documents without any prescribed length limit or format.
We benchmark \textit{ECTSum} with state-of-the-art summarizers across various metrics evaluating the content quality and factual consistency of the generated summaries.
Finally, we present a simple-yet-effective approach, \textbf{\model}, to generate a set of bullet points that precisely capture the important facts discussed in the calls.

\end{abstract}

\section{Introduction}
\label{sec:intro}

\textit{Earnings Calls}, typically a teleconference or a webcast, are hosted by publicly traded companies to discuss important aspects of their quarterly (10-Q), or annual (10-K) earnings reports, along with current trends and future goals that help financial analysts and investors to review their price targets and trade decisions \cite{GIVOLY1980221, Frankel1999, Bowen2002, keith-stent-2019-modeling}. 
The corresponding call transcripts (called \textbf{Earnings Call Transcripts}, abbreviated as \textbf{ECT}s) are typically in the form of long unstructured documents consisting of thousands of words. 
Hence, it requires a great deal of time and effort, even on the part of trained analysts, to quickly summarize the key facts covered in these transcripts. 
Given the importance of these calls, they are often summarized by media houses such as \textit{Reuters} and \textit{BusinessWire}.
The scale of such effort, however, calls for the development of efficient methods to automate this task which in turn necessitates the creation of a benchmark dataset. 

\begin{table}[!t]
    \centering
    \footnotesize
    \begin{tabular}{p{72mm}}
    \toprule
    
    \begin{itemize}[nosep, leftmargin=*]
        \item QUARTERLY EARNINGS PER SHARE \$1.52.
        \item QUARTERLY TOTAL NET SALES \$97.28 BILLION VERSUS \$89.58 BILLION REPORTED LAST YEAR.
        \item BOARD OF DIRECTORS AUTHORIZED AN INCREASE OF \$90 BILLION TO THE EXISTING SHARE REPURCHASE PROGRAM.
        \item QUARTERLY IPHONE REVENUE \$50.57 BILLION VERSUS \$47.94 BILLION REPORTED LAST YEAR.
    \end{itemize}\\
    
    \toprule
    
    \end{tabular}
    \caption{\textbf{ECTSum}: Excerpt from the \textit{Reuters} article\tablefootnote{\href{https://tinyurl.com/yc3z9sbj}{https://tinyurl.com/yc3z9sbj}} corresponding to the ECT\tablefootnote{\href{https://tinyurl.com/uyby3vh4}{https://tinyurl.com/uyby3vh4}} for {\bf Apple Q2 2022}.}
    \vspace{-1em}
    \label{tab:reuters_summary}
\end{table}

Towards this goal, we present \textbf{ECTSum}, a new benchmark dataset for bullet-point summarization of long ECTs. 
As discussed in Section~\ref{subsec:dataset:creation}, first we crawled around 7.4K ECTs from \textit{The Motley Fool}\footnote{\href{https://www.fool.com/earnings-call-transcripts/}{https://www.fool.com/earnings-call-transcripts/}}, posted between January 2019 and April 2022, corresponding to the \textit{Russell 3000 Index} companies\footnote{\href{https://www.investopedia.com/terms/r/russell\_3000.asp}{https://www.investopedia.com/terms/r/russell\_3000.asp}}.
\textit{\textbf{Reuters}} was chosen to be the source of our target summaries, per consultation with domain experts, since the expert-written articles posted on \textit{Reuters} effectively capture the key takeaways from earnings calls. 
However, searching for \textit{Reuters} articles corresponding to the collected ECTs was especially challenging, since the task was non-trivial.
Given the fact that not all calls are tracked, after carefully performing data cleaning and addressing pairing issues, we arrive at a total of \textbf{2,425 document-summary pairs} as part of the dataset.

What makes \textit{ECTSum} truly different from others is the way the summaries are written. 
Instead of containing well-formed sentences, the articles contain \textit{telegram-style bullet-points} precisely capturing the important metrics discussed in the earnings calls.
A sample reference summary from our dataset corresponding to the 2nd quarter 2022 earnings call of \textit{Apple} is shown in Table~\ref{tab:reuters_summary}.
There are several other factors that make \textit{ECTSum} a challenging dataset. 
First, the document-to-summary \textbf{compression ratio of 103.67} is the \textbf{highest} among existing long document summarization datasets with comparable document lengths (Table~\ref{tab:dataset}). 
Hence, in order to do well, trained models need to be highly precise in capturing the most relevant facts discussed in the ECTs in as few words as possible.

Second, existing long document summarization datasets such as Arxiv/PubMed \cite{arxiv_pubmed}, BigPatent \cite{bigpatent}, FNS \cite{fns}, and GovReport \cite{hepos}, have fixed document layouts. 
ECTs, on the other hand, are free-form documents with salient information spread throughout the text (please refer Section \ref{subsec:dataset:stats}).
Hence, models can no longer take advantage of learning any stylistic signals \cite{booksum}. 
Third, the average length of ECTs is around 2.9K words (before tokenization). On the other hand, neural models employing BERT \cite{devlin-etal-2019-bert}, T5 \cite{t5}, or BART \cite{lewis-etal-2020-bart} as document encoders cannot process documents longer than 512/1024 tokens. 
Hence, despite achieving state-of-the-art performances on short-document summarization datasets such as CNN/DM \cite{cnndm}, Newsroom \cite{newsroom}, and XSum \cite{xsum}, etc., such models cannot be readily applied to effectively summarize ECTs.

We benchmark the performance of several representative supervised and unsupervised summarizers on our newly proposed dataset (Section~\ref{subsec:exp:baselines}).
Among supervised methods, we select state-of-the-art extractive, abstractive, and long document summarization approaches.
Given the pattern of source transcripts and target summaries, we then present \textbf{\model}, a simple yet effective pipeline approach for the task of ECT summarization (Section~\ref{sec:method}). 
It consists of an \textbf{extractive summarization} module followed by a \textbf{paraphrasing} module. 
While, the former is trained to identify salient sentences from the source ECT, the latter is trained to paraphrase ECT sentences to short abstractive telegram-style bullet-points that precisely capture the numerical values and facts discussed in the calls.

In order to demonstrate the challenges of the proposed \textit{ECTSum} dataset, competing methods are evaluated on several metrics that assess the \textit{content quality} and \textit{factual consistency} of the model-generated summaries. 
These metrics are discussed in Section~\ref{subsec:exp:metrics}. 
We discuss the comparative results of all considered methods against automatic evaluation metrics in Section~\ref{subsec:exp:results}. 
Given the complex nuances of financial reporting, we further conduct a human evaluation experiment (survey results reported in Section~\ref{subsec:analysis:humaneval}) where we hire a team of financial experts to manually assess and compare the summaries generated by \textit{\model}, and those of our strongest baseline. 
Overall, both automatic and manual evaluation results show \textbf{\model} to outperform strong state-of-the-art baselines, which demonstrates the advantage of a simple approach.



Our contributions can be summarized as follows:
\begin{itemize}[leftmargin=*, nosep]
    \item We present \textbf{ECTSum}, the first long document summarization dataset in the finance domain that requires models to process long unstructured earning call transcripts and summarize them in a few words while capturing crucial metrics and maintaining factual consistency.
    \item We propose \textbf{\model}, a simple approach to effectively summarize ECTs while ensuring factual correctness of the generated content. We establish its better efficacy against strong summarization baselines across all considered metrics evaluating the content quality and factual correctness of model-generated summaries.
    \item Our dataset and codes are publicly available at \href{https://github.com/rajdeep345/ECTSum}{ https://github.com/rajdeep345/ECTSum}
\end{itemize}



\section{Related Works}
\label{sec:related}



Automatic text summarization, \textit{extractive} \cite{nallapati_summaRunner, matchsum}, \textit{abstractive} \cite{zhang2019pegasus, lewis-etal-2020-bart}, as well as \textit{long document summarization} \cite{bigbird, longformer} have seen tremendous progress over the years \cite{huang-etal-2020-achieved}.
Several works also exist on \textit{controllable summarization} \cite{sigir2020, amplayo-etal-2021-aspect} and, in specific domains, such as \textit{disaster} \cite{wsdm2022}, and \textit{legal} \cite{shukla2022legal}.
However, the field of financial data summarization remains largely unexplored, primarily due to the unavailability of suitable datasets.
\citet{passali-etal-2021-towards} have recently compiled a financial news summarization dataset consisting of around 2K \textit{Bloomberg} articles with corresponding human-written summaries. However, similar to other popular \textit{newswire} datasets such as CNN/DM \cite{cnndm}, Newsroom \cite{newsroom}, XSum \cite{xsum}, the documents (news articles) themselves are only a few hundred words long, hence limiting the practical importance of model generated summaries \cite{booksum}.

To the best of our knowledge, \textit{FNS} \cite{fns} is the only available financial summarization dataset, released as part of the \textit{Financial Narrative Summarization Shared Task 2020} \footnote{\href{http://wp.lancs.ac.uk/cfie/fns2020/}{http://wp.lancs.ac.uk/cfie/fns2020/}}. 
In \textit{FNS}, annual reports of UK firms constitute the documents, and a subset of \textit{narrative} sections from the reports are given verbatim as reference summaries.
However, \textit{ECTSum} differs from \textit{FNS} on several accounts. 

First, our target summaries consist of a small set of telegram-style bullet-points, whereas the ones in \textit{FNS} are large extractive portions from respective source documents. 
Second, \textit{ECTSum} has a very high document-to-summary \textit{compression ratio} (refer Section \ref{subsec:dataset:stats}), because of which the models are expected to generate extremely concise summaries of around 50 words from lengthy unstructured ECTs, around 2.9K words long.
In contrast, the expected length of model-generated summaries on \textit{FNS} is around 1000 words.
Finally, the models developed on \textit{FNS} are specifically trained to identify and summarize the \textit{narrative} sections, while completely ignoring others containing facts, and figures that reflect the firm's annual financial performance.
Excluding these key performance indicators from summaries limits their practical utility to stakeholders.
Models trained on \textit{ECTSum}, on the other hand, are specifically expected to capture salient financial metrics such as sales, revenues, current trends, etc. in as few words as possible.

Previously, \citet{ect_sum_unsup} had attempted to summarize earnings calls using standard unsupervised approaches. We are however the first to propose and exhaustively benchmark a large scale financial long document summarization dataset involving earnings call transcripts.

\begin{table*}[!thb]
    \centering
    \resizebox{\linewidth}{!}{
    \begin{tabular}{lcccccc}
        \hline
        \textbf{Dataset} & \textbf{\# Docs.} & \textbf{Coverage} & \textbf{Density} & \textbf{Comp. Ratio} & \multicolumn{2}{c}{\textbf{\# Tokens}} \\
         & & & & & Doc. & Summary \\ 
        \hline
        \textsc{Arxiv/PubMed} \cite{arxiv_pubmed}\textsuperscript{$\ast$} & 346,187 & 0.87 & 3.94 & 31.17 & 5179.22 & 257.44 \\
        \textsc{BillSum} \cite{billsum}\textsuperscript{$\dagger$} & 23,455 & - & 4.12 & 13.64 & 1813.0 & 207.7 \\
        \textsc{BigPatent} \cite{bigpatent}\textsuperscript{$\ast$} & 1,341,362 & 0.86 & 2.38 & 36.84 & 3629.04 & 116.67 \\
        \textsc{GovReport} \cite{hepos}\textsuperscript{$\dagger$} & 19,466 & - & 7.60 & 19.01 & 9409.4 & 553.4 \\
        \textsc{BookSum} Chapters \cite{booksum}\textsuperscript{$\ast$} & 12,293 & 0.78 & 1.69 & 15.97& 5101.88 & 505.32 \\
        \hline
        ECTSum & 2,425 & 0.85 & 2.43 & \textbf{103.67} & 2916.44 & 49.23 \\
        \hline
        
    \end{tabular}
    }
    \caption{Comparing the statistics of ECTSum dataset with existing long document summarization datasets. The numbers for the datasets marked with \textsuperscript{$\ast$} are copied from \citet{booksum}, whereas the ones marked with \textsuperscript{$\dagger$} are copied from \citet{hepos}. Numbers which were not reported are left blank. ECTSum has the highest \textit{compression ratio} among all the datasets while having comparable \textit{coverage} and \textit{density} scores.}
    \vspace{-0.2em}
    \label{tab:dataset}
\end{table*}
\section{Dataset}
\label{sec:dataset}

This section describes our dataset, \textbf{ECTSum}, including the data sources, and the steps taken to sanitize the data, in order to obtain the document-summary pairs. 
Finally, we conduct an in-depth analysis of the dataset and report its statistics.

\subsection{Data Collection}
\label{subsec:dataset:source}

ECTs of listed companies are publicly hosted on \textit{The Motley Fool} \footnote{\href{https://www.fool.com/earnings-call-transcripts/}{https://www.fool.com/earnings-call-transcripts/}}. 
We crawled the web pages corresponding to all available ECTs for the \textit{Russell 3000 Index} companies\footnote{\href{https://www.investopedia.com/terms/r/russell\_3000.asp}{https://www.investopedia.com/terms/r/russell\_3000.asp}} posted between January 2019 and April 2022. 
In the process, we obtain a total of 7,389 ECTs.
The HTML web pages were parsed using the BeautifulSoup\footnote{\href{https://crummy.com/software/BeautifulSoup/}{https://crummy.com/software/BeautifulSoup/}} library. 
ECTs typically consist of two broad sections: \textit{Prepared Remarks}, where the company's financial results, for the given reporting period, are presented; and \textit{Question and Answers}, where call participants ask questions regarding the presented results.
We only consider the unstructured text corresponding to the \textit{Prepared Remarks} section to form the source documents.

Collecting expert-written summaries corresponding to these ECTs was a far more challenging task.
\textit{Reuters}\footnote{\href{https://www.reuters.com/business/}{https://www.reuters.com/business/}} hosts a huge repository of financial news articles from around the world.
Among these, are articles, written by analysts, that summarize earnings calls events in the form of a few bulleted points (see Table~\ref{tab:reuters_summary}).
After manually going through several such articles, and after consulting experts from \textit{Goldman Sachs, India}, we understood that these articles precisely capture the key takeaways\footnote{\href{https://tinyurl.com/27ehcxzf}{https://tinyurl.com/27ehcxzf}} from earnings calls. 
Accordingly, using the company codes and dates of the earnings call events corresponding to the collected ECTs, we crawled \textit{Reuters} web pages to search for relevant articles.
We obtained 3,013 \textit{Reuters} articles in the process.

\subsection{Data Cleaning and Pairing}
\label{subsec:dataset:creation}

\noindent {\bf Cleaning the ECTs:} Almost all earnings calls (and hence the corresponding transcripts) begin with an introduction by the call moderator/operator. 
We remove these statements since they do not relate to the financial results discussed thereafter. 
Some calls directly start with the \textit{Questions and Answers}, in which case we exclude them from the collection. 

\vspace{0.5em}
\noindent {\bf Cleaning the summaries:}
For the \textit{Reuters} (summary) articles, first we performed simple pre-processing to split the text into sentences. 
In many articles, we observed sentences ending with the phrase \textsc{\small REFINITIV IBES DATA}. 
Such sentences report estimates made by \textit{Refinitiv}\footnote{\href{https://tinyurl.com/2p9e6kh2}{https://tinyurl.com/2p9e6kh2}} analysts on the earnings of publicly traded companies. 
We remove these sentences as they \textbf{do not} correspond to the actual results discussed in the earnings calls (as understood from our discussion with financial experts). 
In the process, we make our target summaries factually consistent with the source documents.

\vspace{0.5em}
\noindent {\bf Creating Document-Summary Pairs:} 
In order to automate the process of pairing an ECT with its corresponding \textit{Reuters} article, first we made sure that the article mentions the same company code as the ECT, and second, it is posted either on the same day or at max one day after the earnings event.
Please refer to Section \ref{subsec:appendix:docsummpairs} for more details.
After obtaining the automatically-matched pairs, the authors manually and independently cross-checked 200 randomly selected ECT(document)-\textit{Reuters}(summary) pairs. 
We found all the pairs to be properly matched.
The process thus ensures accuracy at the cost of obtaining a smaller amount of (sanitized) data.
The dataset can however be easily extended as future earnings calls are covered by media houses, such as \textit{Reuters}, and \textit{BusinessWire}.

\begin{figure}
    \centering
	\includegraphics[width=0.9\columnwidth]{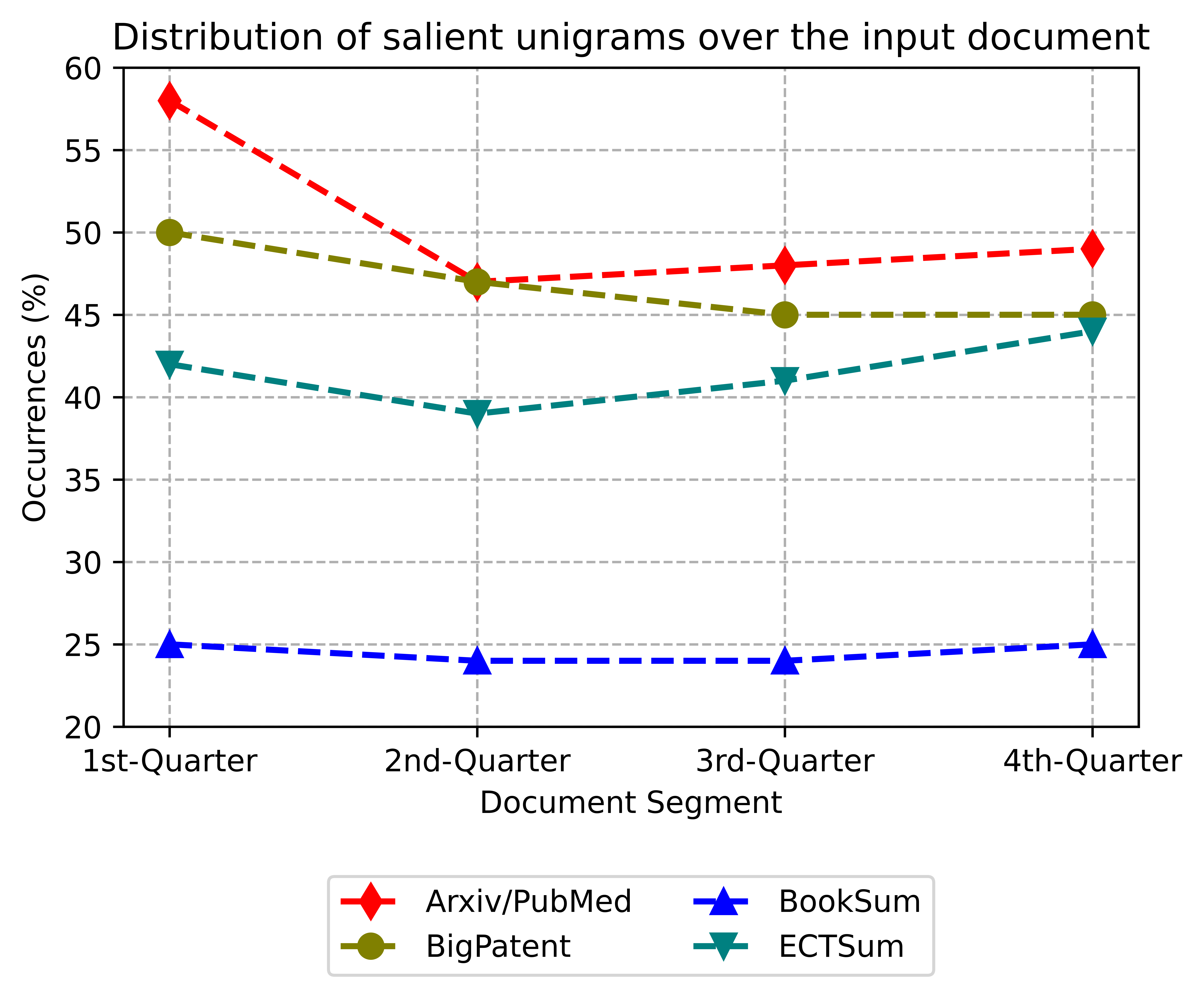}
	\caption{Salient unigram distribution in four equally sized segments of the source text. Higher percentages indicate higher unigram overlap. Percentages more than 25 indicate there are repetitions.
	}
	\vspace{-1em}
	\label{fig:unigram_dist}
\end{figure}

\subsection{Statistics and Analysis}
\label{subsec:dataset:stats}

\begin{figure*}
    \centering
	\includegraphics[width=0.85\textwidth]{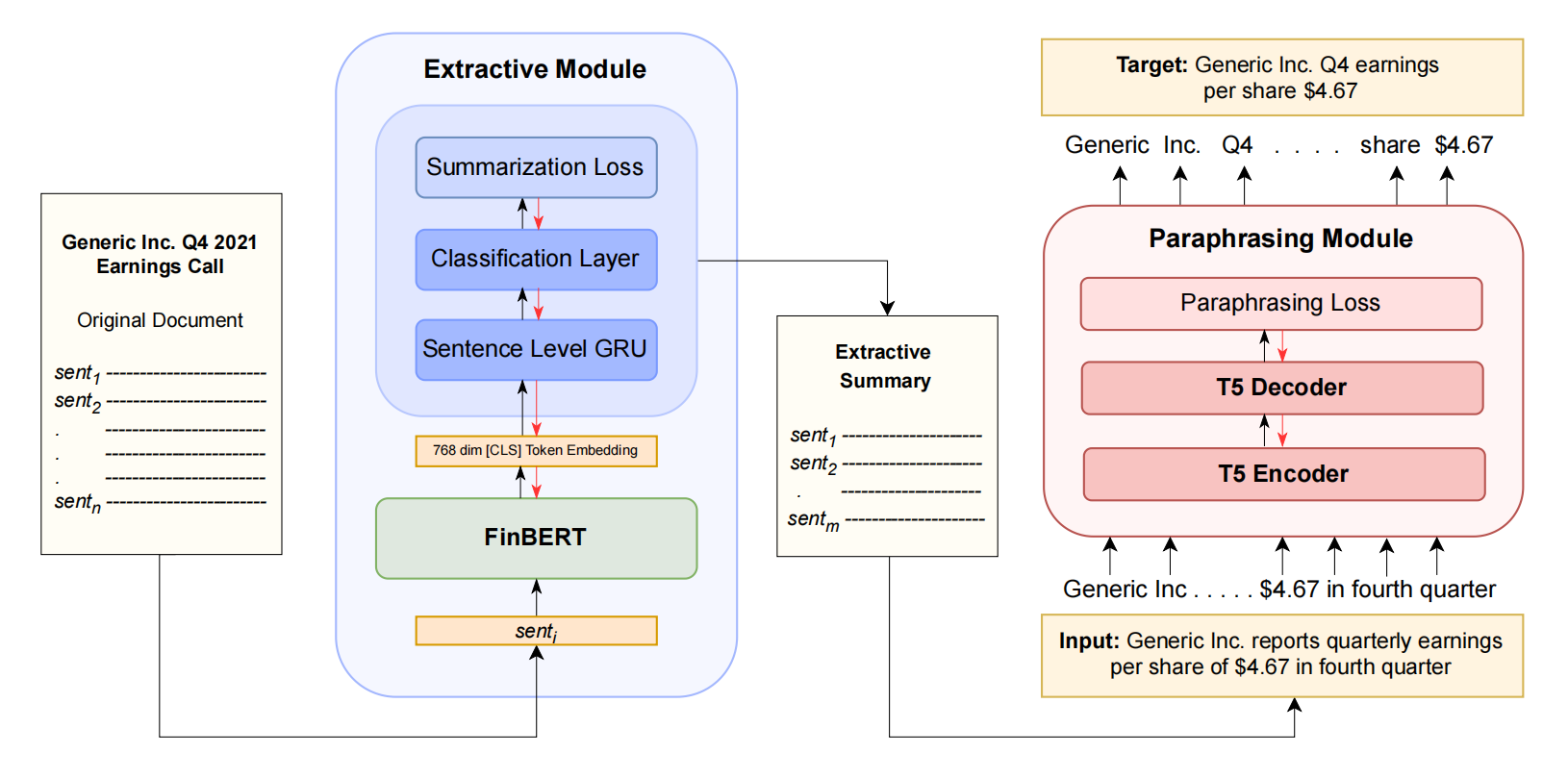}
	\caption{\model: Our Proposed Summarization Framework. It consists of an \textit{Extractive Module} that is trained to select highly salient sentences from the source document. The \textit{Paraphrasing Module} is then trained to paraphrase the ECT sentences to the (\textit{Reuters}) format of target summary sentences.}
	\label{fig:model}
\end{figure*}

The data cleaning and pairing process described above resulted in a total of \textbf{2,425 document-summary pairs}, with average document length of around 2.9K words and average target summary length of around 50 words. 
We randomly split the data to form the train (70\%), validation (10\%) and test (20\%) sets. 
In Table \ref{tab:dataset}, we report various dataset statistics, as defined by \citet{newsroom}, for the \textit{ECTSum} corpus and compare them with the existing long document summarization datasets. 
While \textit{Coverage} quantifies the extent to which a summary is derivative of a text, \textit{Density} measures how well the word sequence of a summary can be described as a series of extractions. Our scores of 0.85 (Coverage) and 2.43 (Density) are fairly comparable with other datasets.
These indicate that although our target summary sentences are short abstractive texts, they are fairly derivable from the ECT content. 
Our document-to-summary \textbf{compression ratio} score of 103.67 is overwhelmingly \textbf{higher than any other dataset}.
This makes \textit{ECTSum} challenging to work on, and requires models to be trained in a way so that they can capture relevant information in as few words as possible. 
Both these factors motivated the design of our proposed approach, \model\ (refer to Section \ref{sec:method}).

Following prior works \cite{hepos, booksum}, we further assess whether the target summary content is confined to certain portions of the source document. 
For this, we plot, in Fig.~\ref{fig:unigram_dist}, the percentage distribution of \textit{salient unigrams} (target summary words excluding stopwords) in four equally sized segments of the source text. 
We observe that the salient content is evenly distributed across all the four segments of the source documents.
This property requires models, trained on \textit{ECTSum}, to process the entire document in order to generate a high quality summary.


\section{The \model \hspace{1pt} Framework}
\label{sec:method}

We observe some important properties of the \textit{Reuters} reference summaries.
They contain a high percentage of word overlap with the source ECT documents. 
However, they are \textit{not} extractive, rather contain a small set of abstractive bullet-points. 
It seems as if the analysts writing these summaries first selected some crucial parts of the ECT, before compressing them into a bullet-point format. 
These properties of the reference summaries motivated us to design a two-stage pipeline approach for summarizing ECTs. 
Our proposed model \textbf{\model} contains two separately trained modules/blocks -- 
(1)~an \textit{Extractive} block that is trained to identify the most relevant sentences from the input ECT document, and
(2)~a \textit{Paraphrasing} block that is trained to rephrase the extracted ECT sentences to the format of target (\textit{Reuters}) sentences, thereby generating a set of bullet points. 
Figure \ref{fig:model} gives an overview of our proposed architecture.

\subsection{The Extractive Module}
\label{subsec:method:ext}

We leverage and suitably modify the architecture of \textit{SummaRuNNer} \cite{nallapati_summaRunner} to design our extractive module. 
The vanilla SummaRuNNer consists of a two-layer bi-directional GRU-RNN. 
The first layer works at the \textit{word-level} to learn contextualized word representations, which are then average-pooled to obtain sentence representations. 
We replace this layer by \textbf{FinBERT} \cite{yang2020finbert}, a BERT model pre-trained on financial communication text, and use it to obtain the individual sentence representations. 
The second layer of bi-directional GRU-RNN works at the \textit{sentence-level} to learn contextualized representations of the input ECT sentences. 
We then obtain the document representation $d$ using the hidden state vectors of sentences from this second layer of bi-directional GRU-RNN as follows:
\vspace{-0.5em}
\begin{equation}
    d = tanh (W_d \frac{1}{N_d} \sum^{N_d}_{i = 1} [h_i^f, h_i^b] + b)
\end{equation}
where $h_i^f$ and $h_i^b$ respectively represent the hidden state vectors of the forward, and backward GRUs corresponding to $s_i$, the $i^{th}$ sentence of the input ECT document. $W_d$ and $b$ represent the weight and bias parameters, respectively. $N_d$ represents the number of  sentences in the document.

Each sentence $s_i$ is sequentially revisited in a second pass where a \textit{classification layer} (Fig. \ref{fig:model}) takes a binary decision regarding its inclusion in the summary as follows:
\begin{equation}\label{eq:summ_decision}
    P (y_i=1) = f(h_i, sum_i, d, p^a_i, p^r_i, \nu_i)
\end{equation}
Here, $h_i$ represents a non-linear transformation of $[h_i^f, h_i^b]$. 
$sum_i$ represents the intermediate representation of the summary formed till $s_i$ is visited.
$p^a_i$, and $p^r_i$ respectively represent the absolute and relative positional embeddings corresponding to $s_i$.
Please refer to \citet{nallapati_summaRunner} for more details.
We add a parameter $\nu_i$ that is set to 1 if $s_i$ contains numerical values, and 0 otherwise. 
Keeping in mind the nature of the target summary sentences, that predominantly discuss metrics and numbers, $\nu_i$ guides the classifier to give higher weightage to sentences containing numerical values. 
Therefore, for each sentence $s_i
$, its \textit{content} $f(h_i)$, \textit{salience} given the document context $f(h_i, d)$, \textit{novelty} considering the summary already formed ($f(h_i, sum_i)$, positional importance, and the fact whether it contains monetary figures, are all taken into account while deciding upon its summary membership.


\subsection{The Paraphrasing Module}
\label{subsec:method:para}

As depicted in Fig. \ref{fig:model}, we fine-tune T5 \cite{t5} to paraphrase the input ECT sentences to the telegram-style (\textit{Reuters}) format of target summary sentences.
During this paraphrasing, special care is taken to ensure that the numerical values in the input sentences are not rephrased wrongly (hallucinated). 
More specifically, during training we replace the numerical values in the input sentences with placeholders such as [num-one], [num-two], etc. 
After obtaining the paraphrased sentences, we replace the placeholders with their original values by performing a simple post-processing step.

\subsection{Training and Inference}
\label{subsec:method:train}

\noindent{\bf Target Summary for Extractive Module.} Corresponding to each sentence (hereby referred to as the `target sentence') in the reference summary (obtained from \textit{Reuters}), first we greedily search for a document sentence (using \textit{regular} expressions) that captures all the numerical values mentioned in the target sentence. 
In case of multiple matches, we select all such document sentences. 
If no match is found, we select the document sentence that is most similar to the target sentence, in terms of cosine similarity between their embeddings obtained using Google's \textit{Universal Sentence Encoder}~\cite{use}. 
The selected set of document sentences serve as the \textit{target summary} for training the \textit{Extractive Module}. 
We train this module by minimizing the \textit{Binary Cross Entropy} loss between the predicted and the true sentence labels.

For training the \textit{Paraphrasing Module}, each sentence in the target summary for the \textit{Extractive Module} becomes the source while the corresponding reference summary sentence becomes the target.
The module is trained by minimizing the \textit{Cross-Entropy} loss between the predicted and target tokens.

During {\bf inference}, a test ECT document is sent as input to the trained \textit{Extractive Module}. 
Sentences corresponding to the extractive summary thus obtained are paraphrased using the trained \textit{Paraphrasing Module} to obtain the final summary.

\section{Experiments and Results}
\label{sec:exp}

In this section, we first enumerate the baselines and evaluation metrics. We then describe our experimental setup, followed by a detailed discussion of our main results. We then report the design and results of a human evaluation experiment conducted to manually assess and compare \model-generated summaries with those of competing baselines. We end this section with a qualitative analysis of model-generated summaries.

\begin{table*}[!thb]
    \centering
    \resizebox{\linewidth}{!}{
    \begin{tabular}{|l|c|c|c|c|c|c|}
        \hline
        \textbf{Model} &
        \textbf{ROUGE-1} & \textbf{ROUGE-2} & \textbf{ROUGE-L} & \textbf{BERTScore} & \textbf{Num-Prec.} & \textbf{SummaC\textsubscript{CONV}} \\
        \hline
        
        \multicolumn{7}{|l|}{\textbf{Unsupervised}} \\
        \hline
        
        LexRank \cite{lexrank} & 0.122 & 0.023 & 0.154 & 0.638 & 1.00 & 1.00 \\
        
        DSDR \cite{dsdr} & 0.164 & 0.042 & 0.200 & 0.662 & 1.00 & 1.00 \\
        
        PacSum \cite{pacsum} & 0.167 & 0.046 & 0.205 & 0.663 & 1.00 & 1.00 \\
        \hline

        \multicolumn{7}{|l|}{\textbf{Extractive}} \\
        \hline
        
        SummaRuNNer \cite{nallapati_summaRunner} & 0.273 & 0.107 & 0.309 & 0.647 & 1.00 & 1.00 \\
        
        BertSumExt \cite{bertsumext} & 0.307 & 0.118 & 0.324 & 0.667 & 1.00 & 1.00 \\
        
        MatchSum \cite{matchsum} & 0.314 & 0.126 & 0.335 & 0.679 & 1.00 & 1.00 \\
        \hline

        \multicolumn{7}{|l|}{\textbf{Abstractive}} \\
        \hline
        
        BART \cite{lewis-etal-2020-bart} & 0.327 & 0.153 & 0.361 & 0.692 & 0.594 & 0.431 \\
        
        Pegasus \cite{zhang2019pegasus} & 0.334 & 0.185 & 0.375 & 0.708 & 0.783 & 0.444 \\
        
        T5 \cite{t5} & 0.363 & 0.209 & 0.413 & 0.728 & 0.796 & 0.508 \\
        \hline

        \multicolumn{7}{|l|}{\textbf{Long Document Summarizers}} \\
        \hline
        
        BigBird \cite{bigbird} & 0.344 & 0.252 & 0.400 & 0.716 & 0.844 & 0.452 \\
        
        
        LongT5 \cite{guo2021longt5} & 0.438 & 0.267 & 0.471 & 0.732 & 0.812 & 0.516 \\
        
        LED \cite{longformer} & 0.450 & 0.271 & 0.498 & 0.737 & 0.679 & 0.439 \\
        \hline

        \multicolumn{7}{|l|}{\textbf{Ours}} \\
        \hline
        
        \model \hspace{1pt} w/o Paraphrasing & 0.313 & 0.137 & 0.351 & 0.714 & 1.00 & 1.00 \\
        
        \model & \textbf{0.467} & \textbf{0.307} & \textbf{0.514} & \textbf{0.764} & \textbf{0.916} & \textbf{0.518} \\
        \hline
        
    \end{tabular}
    }
    \caption{Comparison of representative summarizers against automatic evaluation metrics. Best scores are \textbf{bold}-ed. For \textit{Num-Prec.} and \textit{SummaC\textsubscript{CONV}}, we highlight the best scores among \textit{abstractive} methods (reasons in Section \ref{subsec:exp:results}). \textbf{\model}-generated summaries \textbf{score the highest} on both content quality as well as factual consistency.}
    \vspace{-0.5em}
    \label{tab:main_res}
\end{table*}

\subsection{Baselines}
\label{subsec:exp:baselines}

We evaluate and compare the summarization performance of a wide range of representative algorithms corresponding to various categories on the \textit{ECTSum} corpus. The categories together with their specific algorithms are enumerated below:
\vspace{5pt}
\begin{enumerate}[nosep, leftmargin=*]
    \item {\bf Unsupervised Approaches:} \textbf{LexRank} \cite{lexrank},  \textbf{DSDR} \cite{dsdr}, \textbf{PacSum} \cite{pacsum}. 
    \item {\bf Extractive Approaches: } \textbf{SummaRuNNer} \cite{nallapati_summaRunner}, 
    \textbf{BertSumEXT} \cite{bertsumext}, \textbf{MatchSum} \cite{matchsum}.
    \item {\bf Abstractive Approaches:} \textbf{BART} \cite{lewis-etal-2020-bart}, \textbf{Pegasus} \cite{zhang2019pegasus}, \textbf{T5} \cite{t5}.
    \item {\bf Long Document Summarizers:} \textbf{BigBird} \cite{bigbird}, \textbf{LongT5} \cite{guo2021longt5}, \textbf{Longformer Encoder Decoder (LED)} \cite{longformer}.
\end{enumerate}
For more details, please refer to the {\bf appendix} \ref{subsec:appendix:baselines}.

\subsection{Evaluation Metrics}
\label{subsec:exp:metrics}

\begin{enumerate}[nosep, leftmargin=*]
    \item For evaluating the content quality of model-generated summaries, we consider \textbf{ROUGE} \cite{rouge}, and \textbf{BERTScore}~\cite{bertscore}. We report the F-1 scores corresponding to ROUGE-1, ROUGE-2, and ROUGE-L.
    \item For assessing the factual correctness of the generated summaries, we consider \textbf{SummaC\textsubscript{CONV}} \cite{10.1162/tacl_a_00453}, a recently proposed NLI-based factual inconsistency detection model.
    \item \textbf{Num-Prec.}: Accurate reporting of monetary figures is crucial in the financial domain.
    However, quantity hallucination is a known problem in abstractive summaries~\cite{zhao-etal-2020-reducing}.
    In order to evaluate the correctness of values captured in summaries, especially the abstractive ones, we define \textit{Num-Prec.} as the fraction of numerals/values in the model-generated summaries that appear in the source text. Please refer \ref{subsec:appendix:metrics}.
\end{enumerate}

\subsection{Experimental Setup}
\label{subsec:exp:settings}

As discussed in Section~\ref{subsec:method:train}, we train the two modules of \model \hspace{1pt} separately. For respectively training the \textit{extractive} (and \textit{paraphrasing}) modules, we initialize the FinBERT\footnote{\href{https://huggingface.co/ProsusAI/finbert}{https://huggingface.co/ProsusAI/finbert}} (and T5\footnote{\href{https://huggingface.co/ramsrigouthamg/t5\_paraphraser}{https://huggingface.co/ramsrigouthamg/t5\_paraphraser}}) parameters using pre-trained weights from Huggingface \cite{wolf-etal-2020-transformers}. In the extractive module, all other parameters were set as defined in \citet{nallapati_summaRunner}. The \textit{Extractive} (\textit{Paraphrasing}) module is trained end-to-end with Adam Optimizer with a learning rate of 1e-5 (2e-5) and batch size 8 (16).

Among the baselines, BART\footnote{\href{https://tinyurl.com/26wwaf2e}{https://tinyurl.com/26wwaf2e}} and Pegasus\footnote{\href{https://tinyurl.com/mrwpj8mj}{https://tinyurl.com/mrwpj8mj}} model parameters were initialized with weights pre-trained on financial data. For others, the \textit{base} version of their respective models were used to initialize the parameter weights. All other model hyperparameters were initialized with default values as specified in the respective papers. 

All models, including the \model \hspace{1pt} modules, were trained end-to-end with hyperparameters fine-tuned on the validation set (recall that we used a 70:10:20 ratio as train:validation:test split).
In each case, the model with the lowest validation loss was used to evaluate the test set. All experiments were performed on a Tesla P100-PCIE (16GB) GPU. BART (1024), Pegasus (512), and T5 (512) have limitations on the length of input text that they can process. Since ECTs contain around 2.9K words on an average, for training these \textit{abstractive} methods, we divided the source documents into multiple chunks, each with length less than or equal to their respective \texttt{max\_token\_len}. Corresponding target summaries were made by selecting a subset of all target summary sentences that were entailed by the sentences in the document chunk under consideration. During inference, a small summary (max 32 tokens) was generated from each document chunk. The unique sentences from all such short summaries were concatenated to produce the overall summary for the entire document. Our \textit{ECTSum} dataset, and codes, including baselines, are publicly available on our \textit{GitHub}\footnote{\href{https://github.com/rajdeep345/ECTSum}{https://github.com/rajdeep345/ECTSum}} repository.

\begin{figure*}
    \centering
    \begin{tabular}{cc}
	\includegraphics[width=0.47\textwidth, height=0.3\textwidth]{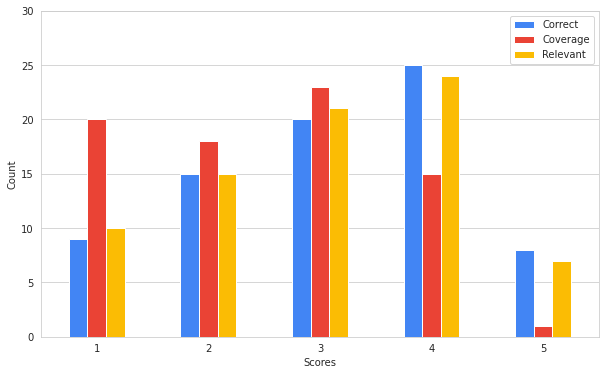} &
	\includegraphics[width=0.47\textwidth, height=0.3\textwidth]{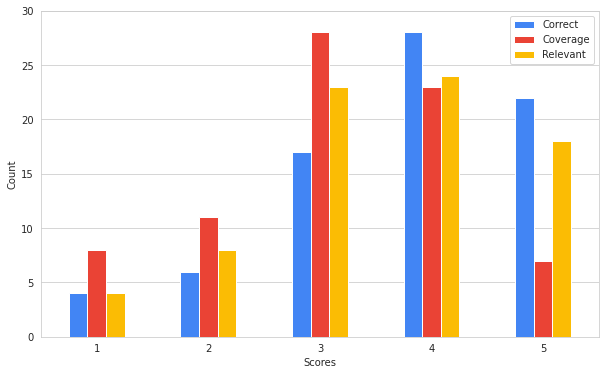} \\
	\textbf{(a) LED summaries} & \textbf{(b) \model \hspace{1pt} summaries} \\
	\end{tabular}
	\caption{Histogram distribution for human evaluation scores assigned to model-generated summaries.}
	\vspace{-0.5em}
	\label{fig:led_vs_our}
\end{figure*}

\subsection{Main Results}
\label{subsec:exp:results}

Table~\ref{tab:main_res} reports the performance of all competing methods on the test set.
All the \textit{unsupervised} methods perform poorly, thereby highlighting the domain-specific nature of the ECT summarization task, and hence the need for supervised training. 
Among the supervised \textit{extractive} methods, \textit{MatchSum}, a state-of-the-art extractive summarizer, has the best scores across all metrics. 
Here, we would like to highlight the advantage of the modifications we made to the vanilla \textit{SummaRuNNer} code. 
Our \textit{Extractive Module}, \model \hspace{1pt} w/o Paraphrasing, when compared to \textit{SummaRuNNer}, achieves 18.7\% improvement on average across all the \textit{ROUGE} scores, and 10.4\% improvement in \textit{BERTScore}. 
This also makes our \textit{Extractive Module} the best performing extractive method across all metrics.
Please note that the \textit{Num-Prec.} and \textit{SummaC\textsubscript{CONV}} scores for all extractive summarizers are always 1.00 because the summary sentences are taken verbatim from the source documents.

Among the \textit{abstractive} methods, \textit{Pegasus} and \textit{BART}, despite being initialized with weights pre-trained on financial data, could not match the performance of \textit{T5}.
Interestingly, both \textit{T5} (0.508) as well its long version, \textit{LongT5} (0.516), have very good factual consistency scores. 
These observations led us to select T5 as the backbone of our \textit{paraphrasing} module. 
\textit{LED} performs better on token overlap metrics (Rouge and BERTscore) but has poor factual consistency scores, highlighting the issue of hallucination in abstractive summarizers \cite{halluc1}. 
To conclude, despite the understandably good performance of \textit{long document summarizers} on the ECT summarization task, our simple extract-then-paraphrase approach, \model, establishes the state-of-the-art performance with overall 6.8\% better \textit{ROUGE} scores, 3.67\% better \textit{BERTScore} scores, 8.5\% better \textit{Num-Prec.} scores, and 0.4\% better factual consistency scores over the respective strongest baselines.


\begin{table}[t]
	\centering
	\resizebox{\columnwidth}{!}{
	\begin{tabular}{cccc}
		\toprule
		
		\textbf{Model} & \textbf{Correctness} & \textbf{Relevance} & \textbf{Coverage} \\
		
		\toprule
		
		\multicolumn{4}{c}{\textbf{Summary-level scores (over 75 summaries)}} \\
		
		\midrule
		
		\textbf{LED better} & 18 \textcolor{darkgray}{(24\%)} & 23 \textcolor{darkgray}{(31\%)} & 21  \textcolor{darkgray}{(28\%)} \\
		\textbf{\model{} better} & 45 \textcolor{darkgray}{(60\%)} & 44 \textcolor{darkgray}{(59\%)} & 48 \textcolor{darkgray}{(64\%)} \\
		\textbf{Both equally good} & 12 \textcolor{darkgray}{(16\%)} & 8 \textcolor{darkgray}{(11\%)} & 6 \textcolor{darkgray}{(8\%)} \\

		\midrule
		
		\multicolumn{4}{c}{\textbf{Expert-level Scores (over 10 experts)}} \\
		
		\midrule
		
		\textbf{LED better} & 3 & 3 & 2 \\
		\textbf{\model{} better} & 7 & 7 & 8 \\
		
		\toprule
		
	\end{tabular}
	}
	\caption{Results for the manual evaluation of model-generated summaries by a team of 10 financial experts.}
	\vspace{-1.5em}
	\label{tab:human_eval}
\end{table}

\subsection{Evaluation by Financial Domain Experts}
\label{subsec:analysis:humaneval}

\begin{table*}
	\centering
	\resizebox{0.8\linewidth}{!}{
	\begin{tabular}{|l|c|c|c|}
		\hline
		\multirow{2}{*}{\textbf{Summary}} & \multicolumn{3}{c|}{\textbf{Evaluation Scores}} \\
		\cline{2-4}
		
		& Correct & Relevant & Coverage \\
		\hline
		
		\multicolumn{4}{|l|}{\textbf{LED-Generated}} \\ 
		\hline
		
		q2 revenue rose 27 percent to \$667 million. & \cmark & \cmark & \\
		sees q3 adjusted earnings per share \$12.80 to \textcolor{red}{\textbf{\$13.90}}. & \xmark & \xmark & \\
		qtrly adjusted net income per diluted share \$3.15. & \cmark & \cmark & 1 \\
		sees fy earnings per common share to be in range of \textcolor{red}{\textbf{\$12}} - \$13.00. & \xmark & \xmark & \\
		sees 2021 revenue \$2.74 billion to \$2.791 billion. & \cmark & \cmark & \\
		
		\hline
		
		\multicolumn{4}{|l|}{\textbf{\model-Generated}} \\ 
		\hline
		
		sees q3 adjusted earnings per share \$3.35 to \$3.55. & \cmark & \cmark & \\
		sees fy \textcolor{red}{\sout{\bf adjusted}} earnings per share \$12.80 to \$13.00. & \xmark & \xmark & \\
		sees fy revenue \$2.74 billion to \$2.79 billion. & \cmark & \cmark & 3 \\
		q2 revenue rose 27 percent to \$667 million. & \cmark & \cmark & \\
		
		
		q2 earnings per share \$2.30. & \cmark & \cmark & \\
		
		\hline
	\end{tabular}}
	\caption{Comparing the summaries generated by LED and \model \hspace{1pt} for a given ECT (details in Section \ref{subsec:analysis:qual}). Parts marked in \textcolor{red}{red} are wrongly generated. \model \hspace{1pt} better preserves the correctness of generated numbers.}
	\vspace{-0.5em}
	\label{tab:qual_ex}
\end{table*}

Given the complex nuances of the financial domain, we get the model-generated summaries evaluated by a team 10 analysts/experts working with \textit{Goldman Sachs Data Science and Machine Learning Group, India} who were well-versed with the concepts of financial reporting, earnings calls, etc.
For this, we create a survey with 75 randomly chosen test set ECTs and their corresponding summaries generated by \model\ and \textit{LED}, our strongest baseline.
Each survey form (please refer to an example\footnote{\href{https://forms.gle/pWtexZqM9TXGGoCAA}{https://forms.gle/pWtexZqM9TXGGoCAA}})  was divided into 5 sections. 
In each section, the participants were required to go through an entire ECT (link provided), and evaluate the two summaries (randomly placed, identity not revealed) on three quality metrics -- \textit{\textbf{factual correctness}, \textbf{relevance}} and \textit{\textbf{coverage}} as defined below:
\begin{itemize}[nosep]
    \item \textbf{Factual Correctness}: For each summary sentence, the task was to asses if it can be supported by the source ECT. 
    \item \textbf{Relevance}: For each summary sentence, the task was to asses if it captures pertinent information relative to the ECT. 
    \end{itemize}
\vspace{0.2em}
The final \textit{correctness}/\textit{relevance} score of the summary is then determined based on the percentage of sentences that are factually correct/relevant as follows: 5 (>80\%), 4 (>60\% \& $\le 80\%$), 3 (>40\% \& $\le 60\%$), 2 (>20\% \& $\le 40\%$), 1 ($\le$20\%).
It is to be noted here that \textit{factual correctness} is an objective metric, whereas \textit{relevance} is a subjective metric. For \textbf{Coverage}, the participants were instructed to assign a score to the overall summary (on a \textit{Likert} scale of 1-5) based upon their impression about the amount/coverage of relevant content present in it.

Participants were adequately remunerated for their involvement in the task. 
The summary of results obtained from this survey are presented in Table~\ref{tab:human_eval}. 
At a summary/sample level, respectively for 60\% (45/75) and 59\% (44/75) of the cases, the summaries generated by \model \hspace{1pt} were found to contain more number of factually correct and relevant sentences than the corresponding LED-generated summaries. 
For 16\% and 11\% of the cases respectively, the scores for \textit{correctness} and \textit{relevance} were the same for both models. 
Also, 64\% of the times, the participants found \model-generated summaries to have a broader \textit{coverage}. 
When we checked the results of individual experts, 70\% of the participants (7 out of 10) found \model-generated summaries to be better with respect to \textit{correctness}, and \textit{relevance}. 
On the other hand, 8 out of 10 participants found \model-generated summaries to have a broader \textit{coverage}. 

The distribution of absolute scores assigned to the summaries are shown in Fig. \ref{fig:led_vs_our} as a histogram plot.
Here again we find that \model-generated summaries are majorly scored $\ge$ 3 across all three metrics, whereas the majority of LED summaries are scored $\le$ 3.
Overall, the  survey results were found to be comprehensively in favor of \model. 

\subsection{Qualitative Analysis}
\label{subsec:analysis:qual}

In Table \ref{tab:qual_ex}, we qualitatively compare the summaries generated by LED and \model \hspace{1pt} corresponding to the earnings call transcript for FleetCor Technologies Inc Q2 2021.\footnote{\href{https://tinyurl.com/mph93w46}{https://tinyurl.com/mph93w46}} 
The expert evaluation scores corresponding to this pair are also reported. 
We observe that LED wrongly produces a few monetary values which make the corresponding sentences factually incorrect. 
Whereas, \model{} maintains the correctness of generated numbers.
This may be attributed to our strategy of replacing numbers with placeholders while training the \textit{Paraphrasing} module (please refer to Section~\ref{subsec:method:para} for details). 
\model{} however makes a factual error in the second sentence where it misses the word \textit{adjusted}. 
In the finance domain \textit{adjusted earnings per share} is different from \textit{earnings per share}. 
These nuances necessitates further research on the ECTSum corpus, and financial summarization in general.

\section{Conclusion}
\label{sec:conclusion}

To our knowledge, \textbf{ECTSum} is the first large-scale long document summarization dataset in the finance domain.
Our documents consist of free-form lengthy transcripts of company earnings calls.
Target summaries consist of a set of telegram-style bullet points derived from corresponding \textit{Reuters} articles that cover the calls.
Drawing observations from the nature of source transcripts and target summaries, we also propose a simple, yet effective \textit{extract-then-paraphrase} approach, \textbf{\model}, that establishes state-of-the-art performance over strong summarization baselines across several metrics.

\textit{ECTSum} is an extremely challenging dataset given the high document-to-summary compression ratio.
Moreover, it is highly extendable as future earnings calls are covered by media houses, such as \textit{Reuters}, \textit{BusinessWire}, etc.
Finally, it is a very specialized one which would otherwise have costed a lot of time and resources if one had to hire experts to write the reference summaries. 
The mere observation that these summaries are created by (expert) analysts and can be leveraged automatically is a major milestone of the paper.
We believe our contributions to the dataset and methodology will attract future research in the finance domain.


\section*{Limitations}
\label{sec:limitations}

In this work, we have restricted ourselves to collecting reference summaries, corresponding to an earnings call transcript, from a single data source, \textit{Reuters}, in our case. Articles summarizing the earnings calls are however published on other media websites as well, for example \textit{CNBC}. In future, we can enrich both the quantity as well as the quality of the dataset, in a scalable manner, by collecting more than one articles from multiple sources thereby resulting in multiple reference summaries corresponding to a single source document.

Despite performing substantially better than strong baseline summarization algorithms, our proposed model \model\ is still a pipeline approach. To our advantage, the improvement in scores over the baselines probably overcomes the increase in the number of model parameters by following an extract-then-paraphrase approach. In future, we would definitely like to address this shortcoming by designing a unified model.

The factual consistency scores obtained using \textit{SummaC\textsubscript{CONV}} are generally low across all methods. This gives us ample scope of improvement which in turn calls for further investigation into the nature of factual errors being made by various approaches. A deeper analysis of dataset nuances can also lead us to interesting ideas to improve performance.

\section*{Ethics Statement}
\label{sec:ethics}
Given the impact of our proposed contributions on the financial community in particular, and wider research community in general, our dataset and codes have been publicly released. 
Our document-summary pairs are derived from public/open domain. 
Still, we may ask users, intending to access our data, to provide a self declaration that the data is to be used solely for research purposes.

\bibliography{main}
\bibliographystyle{acl_natbib}

\clearpage
\appendix

\section{Appendix}

\subsection{Creating Document-Summary Pairs}
\label{subsec:appendix:docsummpairs}
In order to automate the process of pairing an ECT with its corresponding \textit{Reuters} article, first we made sure that the article mentions the same company code as the ECT, and second, it is posted either on the same day or at max one day after the earnings event.
In some cases, we found multiple articles for the same ECT.
Upon manual inspection, we classified them into two broad categories: (1)~(multiple) articles summarizing the same earnings call, but in parts; (2)~articles covering news not directly related to the earnings call. 
We took the articles of the first category, and merged their distinct sentences into one summary file.
After obtaining the automatically-matched pairs, the authors manually and independently cross-checked 200 randomly selected ECT(document)-\textit{Reuters}(summary) pairs. 
We found all the pairs to be properly matched.
The process thus ensures accuracy at the cost of obtaining a smaller amount of (sanitized) data.
The dataset can however be easily extended as future earnings call events are covered by media houses, such as \textit{Reuters}, \textit{CNBC}, and \textit{BusinessWire}. We also propose to release subsequent versions of \textit{ECTSum} on our \textit{Github}\footnote{\href{https://github.com/rajdeep345/ECTSum}{https://github.com/rajdeep345/ECTSum}} repository.

\subsection{Baselines}
\label{subsec:appendix:baselines}

We evaluate and compare the summarization performance of a wide range of representative algorithms on the \textit{ECTSum} corpus as briefed below:

\subsubsection{Unsupervised Approaches}

\begin{itemize}[leftmargin=*]
    \item \textbf{LexRank} \cite{lexrank} uses a graph-based lexical centrality metric to score and summarize the document sentences.
    \item \textbf{DSDR} \cite{dsdr} produces a summary consisting of sentences that can best reconstruct the original document.
    \item \textbf{PacSum} \cite{pacsum} is a graph-based algorithm that redefines sentence centrality by taking into account their relative positions in the document to build a directed graph to be used for document summarization.
\end{itemize}

\subsubsection{Extractive Approaches}

\begin{itemize}[leftmargin=*]
    \item \textbf{SummaRuNNer} \cite{nallapati_summaRunner}: Vanilla version of our \textit{Extractive Module} (Section \ref{subsec:method:ext}).
    \item \textbf{BertSumEXT} \cite{bertsumext} takes pre-trained BERT \cite{devlin-etal-2019-bert} as the sentence encoder and an additional Transformer as the document encoder. A classifier on sentence representations is used for sentence selection.
    \item \textbf{MatchSum} \cite{matchsum} generates a set of candidate summaries from the output of \textit{BertSumEXT}. The candidate that matches best with the document is considered as the final summary.
\end{itemize}

\subsubsection{Abstractive Approaches}

\begin{itemize}[leftmargin=*]
    \item \textbf{BART} \cite{lewis-etal-2020-bart} introduces a denoising autoencoder for pre-training sequence to sequence tasks including summarization.
    \item \textbf{Pegasus} \cite{zhang2019pegasus} introduces a novel pre-training strategy, \textit{Gap Sentence Generation}, especially suitable for abstractive summarization.
    \item \textbf{T5} \cite{t5} systematically applies transfer learning techniques for seq-to-seq generation tasks, including summarization.
\end{itemize}

\subsubsection{Long Document Summarizers}

\begin{itemize}[leftmargin=*]
    \item \textbf{BigBird} \cite{bigbird} applies sparse, global, and random attentions to overcome the quadratic dependency of BERT while preserving the properties of full-attention models. Consequently, it can handle longer context.
    \item \textbf{LongT5} \cite{guo2021longt5} extends the original T5 encoder with \textit{Transient Global} attentions to handle long inputs. The model is pre-trained using the PEGASUS strategy.
    \item \textbf{Longformer Encoder Decoder (LED)} \cite{longformer} is a \textit{Longformer} variant for supporting long document generative seq-to-seq tasks. It uses an attention mechanism that scales linearly with sequence length, making it easy to process documents of thousands of tokens or longer.
\end{itemize}

\subsection{Evaluation Metrics}
\label{subsec:appendix:metrics}

For evaluating the content quality and factual correctness of the model-generated summaries, we consider the following evaluation metrics:

\vspace{5pt}
\begin{itemize}[leftmargin=*]
    \item \textbf{ROUGE} \cite{rouge} measures the textual overlap (n-grams, word sequences) between the generated summary and the reference summary. In this work, we report the F-1 scores corresponding to ROUGE-1, ROUGE-2, and ROUGE-L.
    
    \item \textbf{BERTScore}~\cite{bertscore} aligns the generated and target summaries on a token-level and uses BERT to compute their similarity scores. It correlates better with human judgements. We installed the latest version (0.3.11) of BERTScore from its official implementation\footnote{\href{https://github.com/Tiiiger/bert\_score}{https://github.com/Tiiiger/bert\_score}}, and calculated the scores with the recommended NLI model \textsc{microsoft/deberta-xlarge-mnli}.
    
    \item \textbf{Num-Prec.}: Accurate reporting of facts and monetary figures is crucial in the financial domain. Extractive summaries are always expected to contain values that appear in the source text. However, quantity/numeral hallucination is a known problem in abstractive summaries, which prior works~\cite{zhao-etal-2020-reducing} have attempted to reduce. Here, we define \textit{Num-Prec.} as the fraction of numerals/values in the model-generated summaries that are consistent with the source text. We use this metric to specifically evaluate the precision/correctness with which abstractive summarizers generate values.
    
    \item \textbf{SummaC\textsubscript{CONV}} \cite{10.1162/tacl_a_00453} is a recently proposed NLI-based factual inconsistency detection model based on aggregation of sentence-level entailment scores for each pair of input document and summary sentences. We used the official implementation\footnote{\href{https://github.com/tingofurro/summac}{https://github.com/tingofurro/summac}} of \textit{SummaC} to obtain the scores for all model-generated summaries.
\end{itemize}


\end{document}